\documentclass[10pt,conference]{IEEEtran}
\IEEEoverridecommandlockouts
\usepackage{cite}
\usepackage{amsmath,amssymb,amsfonts}
\usepackage{algorithmic}
\usepackage{graphicx}
\usepackage{textcomp}
\usepackage{xcolor}

\def\BibTeX{{\rm B\kern-.05em{\sc i\kern-.025em b}\kern-.08em
    T\kern-.1667em\lower.7ex\hbox{E}\kern-.125emX}}
\begin{document}

\title{Transferable Physics-Informed Representations via Closed-Form Head Adaptation}


\author{
\IEEEauthorblockN{
Jian Cheng Wong\IEEEauthorrefmark{1}*,
Isaac Yin Chung Lai\IEEEauthorrefmark{2},
Pao-Hsiung Chiu\IEEEauthorrefmark{1},
Chin Chun Ooi\IEEEauthorrefmark{1},
Abhishek Gupta\IEEEauthorrefmark{3},
Yew-Soon Ong\IEEEauthorrefmark{1}\IEEEauthorrefmark{4}
}
\IEEEauthorblockA{\IEEEauthorrefmark{1}Institute of High Performance Computing (IHPC), Agency for Science, Technology and Research (A*STAR), Singapore}
\IEEEauthorblockA{\IEEEauthorrefmark{2}National University of Singapore (NUS), Singapore}
\IEEEauthorblockA{\IEEEauthorrefmark{3}Indian Institute of Technology Goa (IIT Goa), India}
\IEEEauthorblockA{\IEEEauthorrefmark{4}Nanyang Technological University (NTU), Singapore}
\thanks{*Corresponding author: wongj@a-star.edu.sg}
}

\maketitle

\begin{abstract}
Physics-informed neural networks (PINNs) have garnered significant interest for their potential in solving partial differential equations (PDEs) that govern a wide range of physical phenomena. By incorporating physical laws into the learning process, PINN models have demonstrated the ability to learn physical outcomes reasonably well. However, current PINN approaches struggle to predict or solve new PDEs effectively when there is a lack of training examples, indicating they do not generalize well to unseen problem instances. In this paper, we present a transferable learning approach for PINNs premised on a fast Pseudoinverse PINN framework (Pi-PINN). Pi-PINN learns a transferable physics-informed representation in a shared embedding space and enables rapid solving of both known and unknown PDE instances via closed-form head adaptation using a least-squares-optimal pseudoinverse under PDE constraints. We further investigate the synergies between data-driven multi-task learning loss and physics-informed loss, providing insights into the design of more performant PINNs. We demonstrate the effectiveness of Pi-PINN on various PDE problems, including Poisson’s equation, Helmholtz equation, and Burgers’ equation, achieving fast and accurate physics-informed solutions without requiring any data for unseen instances. Pi-PINN can produce predictions 100--1000 times faster than a typical PINN, while producing predictions with 10--100 times lower relative error than a typical data-driven model even with only two training samples. Overall, our findings highlight the potential of transferable representations with closed-form head adaptation to enhance the efficiency and generalization of PINNs across PDE families and scientific and engineering applications.
\end{abstract}

\begin{IEEEkeywords}
physics-informed neural networks, transfer learning, multi-task learning, generalization
\end{IEEEkeywords}

\section{Introduction}
Physics-informed neural networks (PINNs) have attracted significant attention as a neural approach for solving partial differential equations (PDEs) that govern a wide range of physical phenomena across science and engineering~\cite{raissi2019physics, karniadakis2021physics}. PINNs incorporate governing laws directly into the learning objective by treating PDE residuals, and associated boundary conditions (BCs) and initial conditions (ICs), as training constraints, resulting in a ``physics-informed'' loss~\cite{wandel2022spline,Kim2021dpm,Kang2023pixel,Yang2023dmis}.

Despite their promise, two practical limitations have repeatedly emerged. First, PINNs often train slowly and can exhibit poor optimization behavior compared to purely data-driven networks, largely due to the stiffness and complex loss landscapes introduced by physics-informed constraints~\cite{wong2022learning, krishnapriyan2021characterizing, wang2021understanding}. Second, even after expensive training, standard PINNs generalize poorly to new PDE instances (e.g., new coefficients, source terms, BCs/ICs, or parameter regimes), limiting their usefulness when extrapolation or rapid redeployment is needed~\cite{cuomo2022scientific}. As a result, PINNs are still commonly used in a single-instance setting where the physics-informed loss is enforced directly for the target PDE, while extending a trained model to new instances typically requires substantial re-training and re-tuning.

In many real-world settings, however, one does not start from scratch: there may exist a small collection of related PDE instances with solutions (or partial labels) that can be leveraged to learn representations that transfer. This motivates a transfer-learning perspective for PINNs, with the objective to build PINN models that are more reusable and adaptable across PDE families and parameter regimes.

Hence, in this paper, we present Pi-PINN, a fast Pseudoinverse PINN framework that learns \textit{transferable physics-informed representations} and enables \textit{closed-form head adaptation} via a least-squares-optimal pseudoinverse under PDE constraints. The core idea is to decouple learning into (i) a shared embedding that captures transferable structure across related PDE instances and (ii) a task-specific output head that can be adapted efficiently through a closed-form linear solve, avoiding costly gradient-based re-optimization for each new instance. In addition, we study how combining data-driven multi-task learning losses with physics-informed residuals can produce more performant and transferable PINN models.

\begin{figure*}[ht!]
    \centering
    \includegraphics[width=.95\textwidth,keepaspectratio]{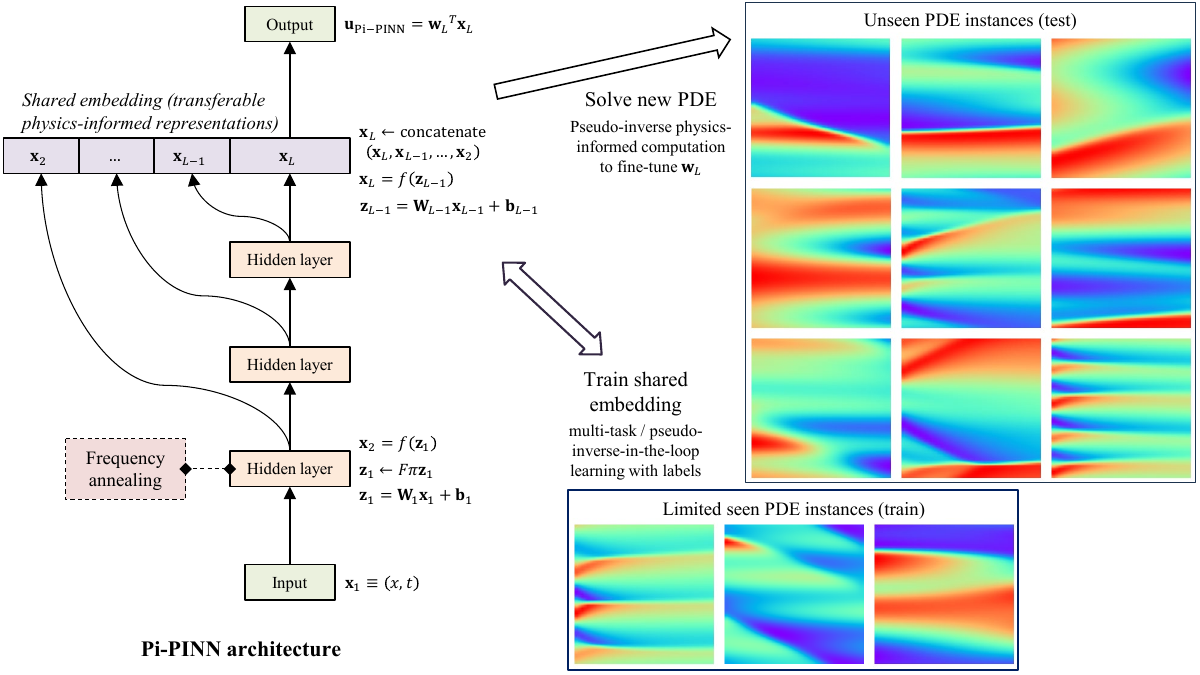}
    \vspace{-0.2cm}
    \caption{The Schematic diagram of the fast Pseudoinverse PINN framework, with the example PDE instances (Burgers' equation).}
    \label{fig:schematic}
\end{figure*}

The key contribution of our work are: (1) we introduce Pi-PINN, a pseudoinverse-based physics-informed learning framework that supports closed-form, least-squares-optimal head adaptation under PDE constraints, substantially reducing the computational cost of adapting PINNs to new PDE instances (see schematic in Figure~\ref{fig:schematic}). (2) We propose a representation-learning formulation for PINNs that learns transferable deep embeddings from related PDE instances, improving generalization across PDE families and parameter regimes. (3) We analyze the synergy between data-driven multi-task learning objectives and physics-informed residual losses, providing practical insights for training PINNs that are both accurate and reusable. (4) Empirically, across various PDE problems including Poisson, Helmholtz, and Burgers equations, Pi-PINN achieves 100--1000$\times$ faster prediction/adaptation than conventional PINNs, while obtaining 10--100$\times$ lower relative error than typical data-driven models in sparse-data regimes (e.g., 2--4 training samples).

\section{Pseudoinverse PINN}

Without loss of generality, we consider the use of PINNs to model the outcomes $u$ in any spatial-temporal domain $x\in\Omega, t\in(0,T]$ by satisfying the following governing equations:
\begin{subequations} \label{eq:pde_ibc_eqn}
    \begin{align}
        & \text{PDE:} & \mathcal{N}[u(x,t)] &= h(x,t), & x\in\Omega, t\in(0,T] \label{eq:pde_eqn} \\
        & \text{BC:} & \mathcal{B}[u(x,t)] &= g(x,t), & x\in\partial\Omega, t\in(0,T] \label{eq:bc_eqn} \\
        & \text{IC:} & u(x,t=0) &= u_0(x), & x\in\Omega \label{eq:ic_eqn}
    \end{align}
\end{subequations} 
where $\mathcal{N}[u(x,t)]$ is a general differential operator which can include linear and nonlinear combinations of temporal and spatial derivatives, and $h(x,t)$ is an arbitrary source term. Eq. \ref{eq:bc_eqn} and Eq. \ref{eq:ic_eqn} specify the BC and IC of the physical system, where $\mathcal{B}[u(x,t)]$ is also a differential operator.

To simplify the notation, we consider that the variation across different PDE instances (including variations in BC and IC) can be parameterized by a set of task parameters $\vartheta$. Every new PDE instance leads to different physical outcomes, $u_\vartheta$, spanning the entire spatial-temporal domain $x\in\Omega, t\in(0,T]$.

A PINN model with sufficient capacity (number of nodes $n_{\text{node}}$, hidden layers $L$) can effectively learn a good approximation to the target output, $u_{\text{PINN}}(x, t; \boldsymbol{w})$, by minimizing the PINN loss function w.r.t. its network parameters $\boldsymbol{w}=(\boldsymbol{W}_1, \mathbf{b}_1, ..., \boldsymbol{W}_{L-1}, \mathbf{b}_{L-1},\mathbf{w}_L)$.

\subsection{Pseudoinverse Physics-Informed $[Pi]^2$ Computation.}

There has been some prior work on transfer learning in the PINN context~\cite{wong2026evopinn,wong2021can, goswami2020transfer}. These typically involve the transfer of network weights from a similar, previously-solved physics scenario as a better initialization for PINN training on a new task, i.e.,  minimizing the physics-informed loss by penalizing any violation of the PDE, BC/IC constraints made by the PINN model outputs on the target PDE. Generally, only the final few layers of the networks are fine-tuned in transfer learning, facilitating the exchange of valuable features acquired when training the PINN model for the previous task.

In the extreme, we can transfer the entire learned representation up to the final nonlinear hidden layer, i.e., $\mathbf{x}_L$, and only fine-tune the weight parameters in the linear output layer $\mathbf{w}_L$. In the absence of target labels for a new PDE instance, this can be done by means of physics-informed fine-tuning that purely minimizes violations of the PDE and BC/IC constraints by the PINN output $u_{\text{PINN}} = \mathbf{w}_L^{T} \mathbf{x}_L$ for the target PDE instance.

For linear PDEs, the process of fine-tuning the output weight parameters $\mathbf{w}_L$ reduces to a linear least-squares computation. This can be observed by examining the PDE, BC/IC constraints of Eq. \ref{eq:pde_ibc_eqn}, where the terms $\mathcal{N}[u_{\text{PINN}}(x, t; \boldsymbol{w})] - h(x,t)$, $u_{\text{PINN}}(x, t=0; \boldsymbol{w}) - u_0(x)$, and $\mathcal{B}[u_{\text{PINN}}(x, t; \boldsymbol{w})] - g(x,t)$, are all linear in terms of the output weights $\mathbf{w}_L$, as long as $\mathcal{N}[\cdot]$ and $\mathcal{B}[\cdot]$ comprise only linear differential operators (as-is true for linear PDE systems)~\cite{dong2021local, calabro2021extreme}. These constraints form a system of linear equations and can be efficiently solved utilizing the Moore-Penrose pseudoinverse~\cite{Penrose56}, giving a least-squares-optimal solution to the unknown $\boldsymbol{w}_L$.

Given a target PDE and a set of PDE $(x^{(p)},t^{(p)})$, BC $(x^{(q)},t^{(q)})$, and IC $(x^{(r)},t^{(r)})$ collocation points, the pseudoinverse computation solves the following weighted system of equations:
\begin{align}
    \left[
    \begin{array}{ccc}
        \lambda_{\textit{PDE}}\ \mathcal{N}[\mathbf{x}_L^T(x^{(p)},t^{(p)})] \\ 
        \vdots \\
        \lambda_{\textit{BC}}\ \mathcal{B}[\mathbf{x}_L^T(x^{(q)},t^{(q)})] \\
        \vdots \\
        \lambda_{\textit{IC}}\ \mathbf{x}_L^T(x^{(r)},t^{(r)}) \\ 
        \vdots \\
    \end{array}
    \right] 
        \mathbf{w}_L  
    &=
    \left[
    \begin{array}{c}
        \lambda_{\textit{PDE}}\ h(x^{(p)},t^{(p)})  \\ 
        \vdots  \\ 
        \lambda_{\textit{BC}}\ g(x^{(q)},t^{(q)})  \\ 
        \vdots  \\
        \lambda_{\textit{IC}}\ u_{0}(x^{(r)})  \\ 
        \vdots  \\ 
    \end{array}
    \right] \nonumber \\
    \mathbf{X} \ \mathbf{w}_L &= \mathbf{y} \label{eq:leastsquares} 
\end{align}
where $\mathbf{w}_L$ are the weight parameters in the output layer to be solved, $\mathbf{x}_L$ is the transferable deep embeddings to be learned, and $\lambda_{\textit{PDE}}$, $\lambda_{\textit{BC}}$, and $\lambda_{\textit{IC}}$ are the weighting hyper-parameters. It has been previously reported in PINN literature that the different physics loss terms, such as PDE and BC/IC constraints, can have vastly different magnitudes, necessitating proper reweighting during training~\cite{wang2021eigenvector}. Similarly, during the pseudoinverse physics-informed computation, the relative weighting of PDE and BC/IC constraints can be adjusted. 

Moreover, we can reformulate Eq. \ref{eq:leastsquares} into
\begin{equation}
    [\lambda_{\textit{PI}}\mathbf{I} + \mathbf{X}^T \mathbf{X}] \ \mathbf{w}_L = \mathbf{X}^T \mathbf{y} \label{eq:leastsquares2}
\end{equation}
with the regularization hyper-parameter $\lambda_{\textit{PI}} \geq 0$. This regularization can significantly improve the conditioning of the problem, thereby improving the quality of the obtained solution. Additionally, Eq. \ref{eq:leastsquares2} reduces the matrix size for the inversion, hence it is more efficient to solve.

Importantly, the pseudoinverse computation allows for rapid fine-tuning (in a single, fast computation) of weight parameters $\boldsymbol{w}_L$ in the final layer of the PINN model for a broad range of linear PDE problems, eliminating the need for gradient-based parameter updates over potentially long and complex convergence trajectories as have been previously reported~\cite{krishnapriyan2021characterizing}. It's also important to note that the pseudoinverse PINN approach can be seamlessly extended to handle nonlinear PDEs by using an iterative update process for the linearized version of the nonlinear constraints~\cite{wong2026baldwin} (as demonstrated in the \textbf{Experiments} on the set of nonlinear Burgers' equation).

\subsection{Pseudoinverse PINN Adapted From Data-Driven Neural Network: MLP$+[Pi]^2$}

Given a set of training data $u_{\vartheta_1}, u_{\vartheta_2}, ..., u_{\vartheta_K}$ from $K$ PDE instances $\vartheta_1, \vartheta_2, ..., \vartheta_K$, a natural first approach is to train a purely data-driven neural network, e.g., a multilayer perceptron (MLP), as the baseline model. We can define the mapping from the input variables $(x,t,\vartheta)$ to the output $u_{\text{MLP}}$.

The resulting learned MLP can then be used to make predictions for new PDE instances with a different $\vartheta$. However, if the training data is sparse in $\vartheta$, i.e., $K$ is small, one may expect such predictions to have a very high chance of being inaccurate~\cite{sun2017revisiting}. Naturally, the level of inaccuracy with respect to the actual $K$ depends on the complexity of the underlying functions, as evident in the \textbf{Experiments} section.

Given the proliferation of data-driven models in the past decade, an interesting question is whether we can take the deep embeddings $\mathbf{x}_L$ learned from a data-driven MLP model and adapt it as a PINN model to solve new PDE instances more accurately, i.e., via the Pseudoinverse physics-informed computation. Crucially, this is a significant paradigm shift from the majority of previous related studies which have successfully demonstrated transfer learning purely between PINN models. This is significant as the training of data-driven models can be significantly less computationally expensive than current PINN methods, and we note the existence of some small set of training data in most real-world problems.

We refer to such data-driven-adapted PINN model as MLP$+[Pi]^2$. Although convenient for transferring prior knowledge from an existing data-driven-trained MLP on seen instances (via $\mathbf{x}_L$) to benefit the prediction of unseen PDE instances, the MLP$+[Pi]^2$ model is not the most effective approach for learning physics-informed representations. While the optimal balance of width and depth can vary with problem, it is common practice in deep learning to enhance learning and generalizability by stacking more hidden layers in an MLP model while maintaining a moderate number of nodes in each hidden layer (going deep rather than wide)~\cite{nguyen2020wide}. When such an MLP model is used, there may be insufficient expressivity and/or degrees of freedom in the $\mathbf{x}_L$ for adaptation. Hence, the  MLP$+[Pi]^2$ model may not adequately satisfy the PDE, BC/IC constraints on the target PDE instance even with the best-fit weight parameters $\mathbf{w}_L$, resulting in inferior predictions. This is empirically observed in the comparison study results in the \textbf{Experiments} section.

\section{Deep Embeddings Neural Architecture and Learning Algorithms for Pi-PINN}

The proper design of a neural architecture and learning algorithm for more performant Pi-PINN remains an open question worthy of more critical consideration. We hypothesize that a wide output layer $\mathbf{x}_L$ is critical as this greatly affects the learnability of a shared embedding space for solving new PDE instances, which in turn determines how well the Pi-PINN can effectively be fine-tuned with the weight parameters $\mathbf{w}_L$ in the output layer by the pseudoinverse computation. Effective learning of this shared embedding space is key to the transfer of prior knowledge to improve PINN performance, thereby ensuring predictions with good compliance to stipulated physics (encoded in the PDE, BC/IC constraints).

With the above requirements in mind, we propose a neural architecture that is potentially more appropriate for the Pi-PINN. Figure~\ref{fig:schematic} gives the Pi-PINN schematic. Essentially, we modify the base MLP by utilizing concatenative skip connections such that all the nonlinear hidden layers are concatenated at the output layer:
\begin{subequations}
    \begin{align}
        u_{\text{Pi-PINN}} &= \mathbf{w}_L^{T} \mathbf{x}_L \quad\quad\quad \text{(\textit{Pi-PINN output layer})}  \label{eq:pipinn_output_layer} \\
        \mathbf{x}_L &\gets \text{concatenate}(\mathbf{x}_L, \mathbf{x}_{L-1}, ..., \mathbf{x}_2) \label{eq:se_output_layer} \\
        \mathbf{x}_L &= f(\boldsymbol{z}_{L-1})  \label{eq:se_final_hidden} \\     
        &\vdots \notag \\
        \mathbf{x}_{2} &= f(\boldsymbol{z}_{1}) \quad\quad\quad\quad\quad\quad \text{(\textit{after activation})} \\     
        \boldsymbol{z}_{1} &\gets F \pi \boldsymbol{z}_{1} \quad\quad\quad\quad \text{(\textit{frequency annealing})} \label{eq:frequency_annealing} \\
        \boldsymbol{z}_{1} &= \boldsymbol{W}_1 \mathbf{x}_1 + \mathbf{b}_1 \quad\quad\quad \text{(\textit{1st hidden layer})} \\
        \mathbf{x}_1 &\equiv (x,t) \quad\quad\quad\quad\quad\quad\quad\quad\quad\quad \text{(\textit{input})} \label{eq:se_input}
    \end{align}
\end{subequations} 

This neural architecture design allows us to increase the output layer width by stacking multiple hidden layers while maintaining a moderate number of nodes in each hidden layer, thereby effectively improving the expressivity of the learned shared embedding and consequently, the prediction performance from data-driven-trained and physics-informed pseudoinverse computation. The proposed incorporation of a physics-informed loss computation during prediction also negates the need for explicit specification of the task parameter $\vartheta$ as input to the MLP for interpolation across different PDEs. It also encourages the learning of a generalizable embedding space for both seen and unseen PDE instances.

In addition, we note that this proposed concatenation for a more expressive shared embedding bears similarity to the way one constructs a polynomial basis space such as the monomial basis space. The additional operations at each hidden layer are analogous to the recurrence relations used in generating Chebyshev polynomials, and incorporation and concatenation of more hidden layers in the MLP essentially results in the creation of a larger (albeit finite) and more expressive basis space (with less truncation). This presents an interesting direction for design of even more performant neural architectures for Pi-PINN computations.

To more effectively learn the embedding space for high-frequency features, we use the sine activation for all the nonlinear hidden layers. In particular, we initialize the network with artificial high-frequency features through manipulating the first hidden layer with a factor of $F \pi$ (Eq. \ref{eq:frequency_annealing}), and allowing these frequencies to be naturally ``reduced" during training\textemdash a process we refer to as ``frequency annealing"~\cite{park2021nerfies} .

We propose two learning algorithms to train the shared embedding $\mathbf{x}_L$ (Eq. \ref{eq:se_output_layer}) from a small set of \textit{a priori}-obtained instances, and compare their adaptation performance in the \textbf{Experiments} section.

\subsection{Multi-Task Learning with Multi-Output Model: HYDRA$+[Pi]^2$}

The multi-task learning approach involves simultaneously learning multiple related tasks using a shared representation\textemdash in our case, the deep embeddings $\mathbf{x}_L$ in the Pi-PINN\textemdash for improved learning and generalization. 

During training, $\mathbf{x}_L$ connects directly to as many output heads as there are different PDE instances $u_{\vartheta_1}, u_{\vartheta_2}, ..., u_{\vartheta_K}$ in the training set. Therefore, we can learn a shared embedding $\mathbf{x}_L$ that matches all seen PDE instances by optimizing the network weights $\boldsymbol{w}$ of a multi-output model to minimize the MSE between target and model outputs. It is worth noting that the model outputs for different PDE instances are controlled by different weight parameters in the output heads, thus creating a good synergy with pseudoinverse computation on new PDE instances. In contrast, the typical MLP model forces a shared output layer for different PDE instances. Multi-task learning with a multi-output model architecture has been explored in PINN literature and is called ``Lernaean Hydra"~\cite{zou2023hydra, Desai2021oneshot}. Hence, we follow the naming convention and refer to the first multi-task learning PINN model as HYDRA$+[Pi]^2$.

For predicting or solving new PDE instances, we discard the multi-output heads from the HYDRA$+[Pi]^2$ model and perform pseudoinverse physics-informed computation to find the best-fit weight parameters $\mathbf{w}_L$ on the desired PDE instance.

\subsection{Pseudoinverse-In-The-Loop Learning: PiL-PINN}

While the synergies between data-driven multi-task learning and pseudoinverse computation shed light on the success of the zero-shot PINN model, the HYDRA$+[Pi]^2$ does not ensure that the learned embedding is optimal for pseudoinverse physics-informed computation. We can further improve the adaptation performance of the Pi-PINN model through pseudoinverse-in-the-loop learning (albeit with higher training costs). This learning algorithm explicitly incorporates the pseudoinverse physics-informed computation during the training stage. The aim is to learn a shared deep embeddings $\mathbf{x}_L$ such that the outputs $u_{\vartheta_1}, u_{\vartheta_2}, ..., u_{\vartheta_K}$ can be best approximated with the pseudoinverse physics-informed computation.

We formulate a data-driven loss function $\mathcal{L}(\boldsymbol{w}^{-}, \Phi(\boldsymbol{w}^{-}))$ as the learning objective for PiL-PINN model optimization, where $\boldsymbol{w}^{-}=(\boldsymbol{W}_1, \mathbf{b}_1, ..., \boldsymbol{W}_{L-1}, \mathbf{b}_{L-1})$ includes all the network weights leading to the shared embeddings $\mathbf{x}_L$, and $\Phi(\boldsymbol{w}^{-})$ comes from the pseudoinverse physics-informed computation for determining the best-fit weight parameters $\mathbf{w}_L$. Critically, the loss function is differentiable and gradient-based optimization methods can be applied. Moreover, we can perform mini-batch training by randomly sampling a subset of PDE instances in every training iteration for greater training efficiency. We refer to our pseudoinverse-in-the-loop learned PINN model as PiL-PINN.

\begin{figure*}[ht!]
    \mbox{\hspace{.5cm}{\includegraphics[width=0.35\textwidth,keepaspectratio]{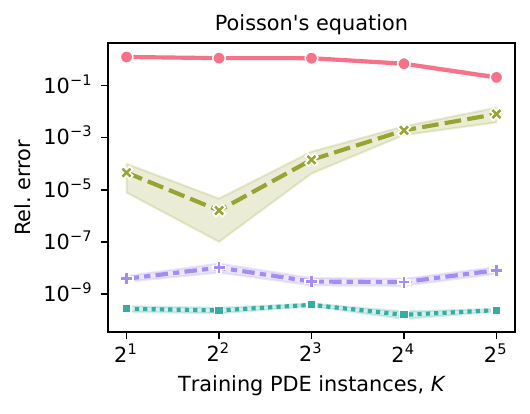}}\hspace{.5cm}{\includegraphics[width=0.35\textwidth,keepaspectratio]{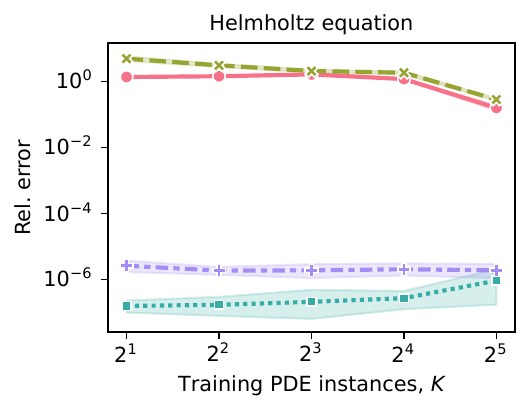}}\hspace{.5cm}{\includegraphics[width=0.18\textwidth,keepaspectratio]{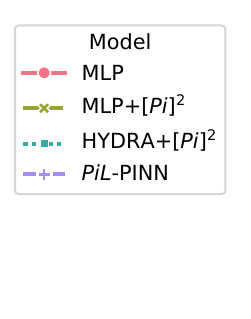}}} \\  
    \mbox{\hspace{.5cm}{\includegraphics[width=0.35\textwidth,keepaspectratio]{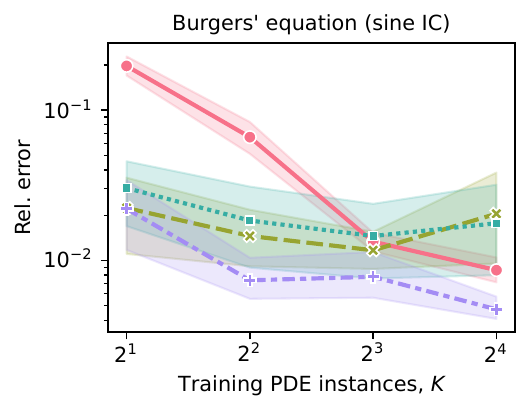}}\hspace{.5cm}{\includegraphics[width=0.35\textwidth,keepaspectratio]{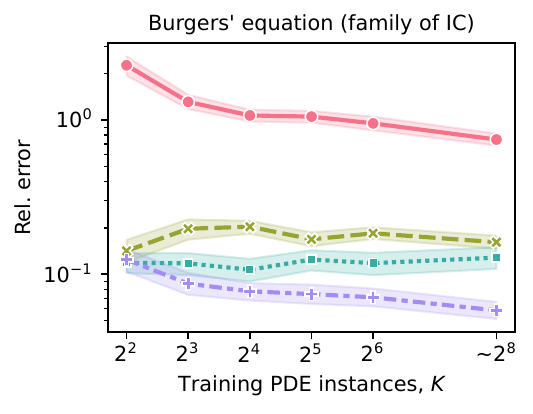}}}
    \vspace{-0.2cm}
    \caption{Comparison of baseline MLP and different Pi-PINN approaches for predicting or solving different PDE problems.}
    \label{fig:main}
\end{figure*}

\section{Experiments}

\subsection{Experimental Setup and Implementation}

We conduct extensive experiments on various real-world PDE problems to show the effectiveness of the different Pi-PINN approaches, i.e., MLP$+[Pi]^2$, HYDRA$+[Pi]^2$, and PiL-PINN. On different problems, we train these models with $K (=2,4,8,...)$ PDE instances with \textit{a priori-provided} solutions, $u_{\vartheta_1}, u_{\vartheta_2}, ..., u_{\vartheta_K}$, and apply the model to predict or solve for remaining PDE instances with no labeled data based on the pseudoinverse physics-informed computation (i.e. zero-shot adaptation). We then compute the relative L2 error on these unseen tasks.

For completeness, we also compare these PINN models with a data-driven MLP, which predicts the unseen PDE instances by interpolating along task parameter $\vartheta$. Below are brief descriptions to the problems.

\subsubsection{Poisson's Equation}
Poisson's equation is a widely studied, representative problem for the broad class of elliptic PDEs. As the solution to potential fields in many physical systems, it can model diverse physics phenomena in electromagnetism and heat and mass transfer (e.g., heat conduction) among many others. In this study, the following 1D Poisson's equation is investigated~\cite{liu2022novel}:
\begin{equation}
    \frac{\partial ^2u}{\partial x^2} = h \quad , x\in[-10,10] \label{eq:poisson}
\end{equation}
with the exact solution $u(x;\omega_1,\omega_2) = \mathrm{sin}(\omega_1 x) + \mathrm{sin}(\omega_2 x) - 0.1 x$. This exact solution is used to derive the corresponding BCs and source term $h(x;\omega_1,\omega_2) = - \omega_1^2\mathrm{sin}(\omega_1 x) - \omega_2^2\mathrm{sin}(\omega_2 x)$. We generate 100 PDE instances with $\omega_1$ and $\omega_2$ uniformly sampled from $(0, 1]$ and $(0, 2]$ respectively. Each PDE instance contains 201 spatial points.

\begin{figure}[ht!]
    \centering
    \includegraphics[width=.35\textwidth,keepaspectratio]{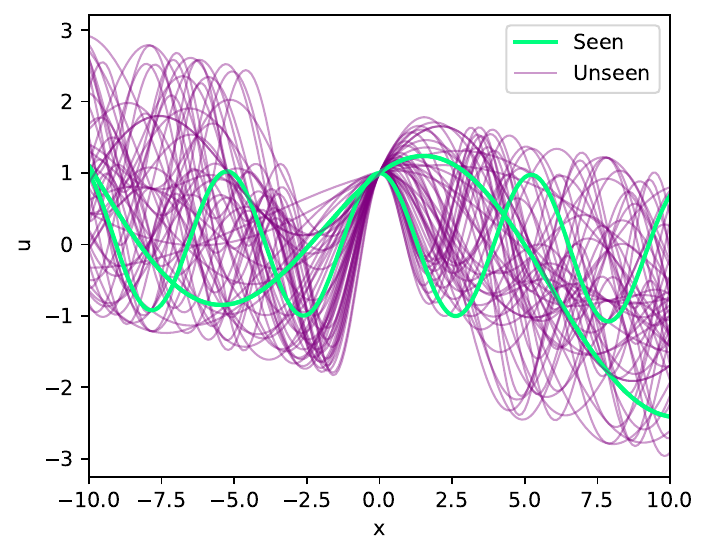}
    \vspace{-0.2cm}
    \caption{The solution of different PDE instances of the Poisson equation problem. Seen examples for model training at $K=2$ is highlighted in green.}
    \label{fig:poisson}
\end{figure}

\begin{figure}[ht!]
    \centering
    \includegraphics[width=.4\textwidth,keepaspectratio]{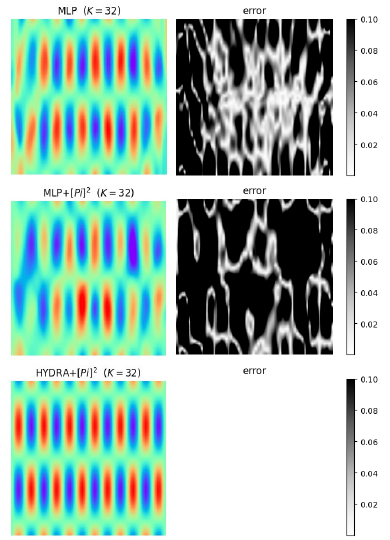}
    \vspace{-0.2cm}
    \caption{The predicted results and corresponding errors for the Helmholtz equation on new PDE instance ($\alpha_1=1.2,\alpha_2=6.0$), for MLP, MLP$+[Pi]^2$, and HYDRA$+[Pi]^2$ trained with $K=32$ PDE instances.}
    \label{fig:helmholtz}
\end{figure}

\subsubsection{Helmholtz Equation}
Helmholtz equation is another well-known model equation that can represent interesting physics such as wave propagation in both acoustics and electromagnetics. In this study, the 2D Helmholtz equation is investigated~\cite{cho2024hypernetwork}:
\begin{equation}
    (\frac{\partial ^2u}{\partial x^2} + \frac{\partial ^2u}{\partial y^2}) + u^2 = h \quad , x\in[-1,1], y\in[-1,1] \label{eq:helmholtz}
\end{equation}
with the exact solution $u(x,y;\alpha_1,\alpha_2) = \left(1 - (\alpha_1 \pi)^2 - (\alpha_2 \pi)^2 \right)\ \mathrm{sin}(\alpha_1 \pi x)\ \mathrm{sin}(\alpha_2 \pi y)$. This exact solution is used to derive the corresponding BCs and source term $h(x,y;\alpha_1,\alpha_2) = (1 - (\alpha_1 \pi)^2 - (\alpha_2 \pi)^2)\  \mathrm{sin}(\alpha_1 \pi x)\ \mathrm{sin}(\alpha_2 \pi y)$. We generated 100 PDE instances with $(\alpha_1$, $\alpha_2)$ uniformly sampled from $(0, 6]$. Each PDE instance contains 64$\times$64 spatial points.

\subsubsection{Burgers' Equation (Sine IC)}
The Burgers' equation is the model equation for a class of transport phenomena of quantities such as momentum, species transport and heat and mass transfer.
We consider the nonlinear Burgers equation as defined by~\cite{liu2022novel, raissi2019physics}:
\begin{equation}
    \frac{\partial u}{\partial t} + u \frac{\partial u}{\partial x} - \gamma \frac{\partial^2u}{\partial x^2} = 0 \quad , x\in[-1,1], t\in(0,1] \label{eq:burgers}
\end{equation}
subject to IC $u(x,t=0) = -\mathrm{sin}(\pi x)$, where the flow solution $u$ represents nonlinear waveform propagation with both compression and rarefaction effects, and $\gamma$ is the viscosity of the fluid. A steep gradient can be observed around $x=0$ after some time $t$. The problem consists of 50 PDE instances with $\gamma$ uniformly sampled from $[0.001, 0.05]$. Each PDE instance contains 129$\times$51 spatial-temporal points.

\begin{figure*}[ht!]
    \centering
    \includegraphics[width=1.\textwidth,keepaspectratio]{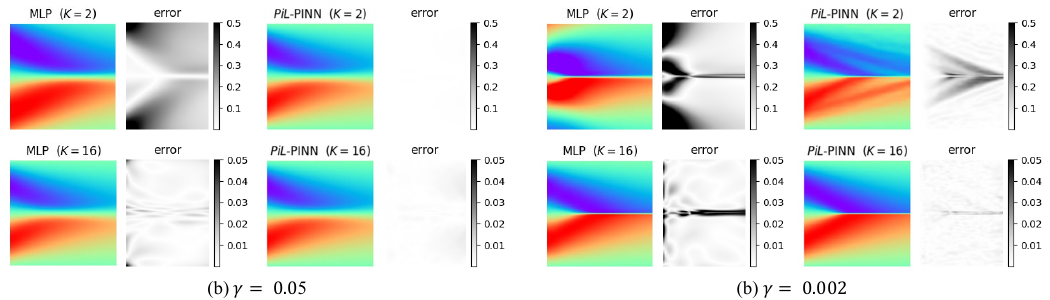}
    \vspace{-0.5cm}
    \caption{The predicted results and corresponding errors for the nonlinear Burger's equation (sine IC), for MLP and PiL-PINN with different PDE instances $K$.}
    \label{fig:sineburger}
\end{figure*}

\begin{figure*}[ht!]
    \centering
    \includegraphics[width=1\textwidth,keepaspectratio]{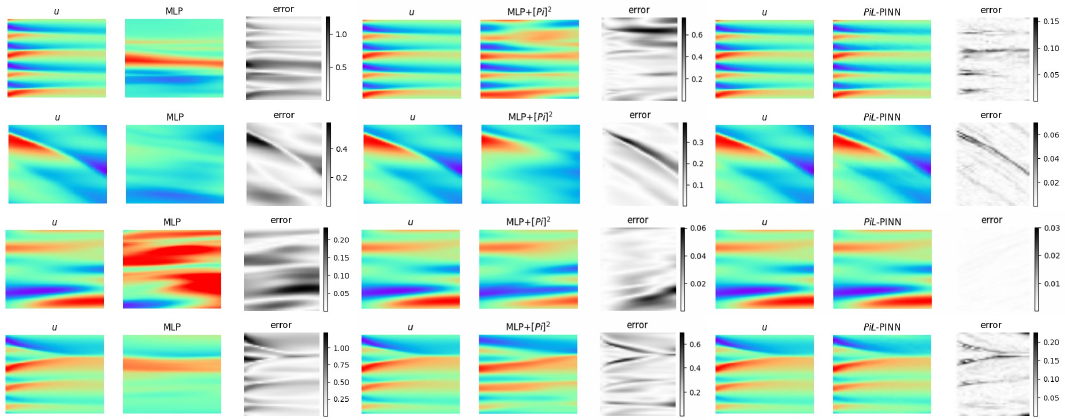}
    \vspace{-0.5cm}
    \caption{The predicted results and corresponding errors for the Burgers' equation (family of IC) on new PDE instances, for MLP, MLP$+[Pi]^2$, and Pil-PINN trained with $K=240$ PDE instances.}
    \label{fig:metaburgers}
\end{figure*}

\subsubsection{Burgers' Equation (Family of IC)}
We further consider the following generalized nonlinear Burgers' equation with a source term,
\begin{equation}
    \frac{\partial u}{\partial t} + u \frac{\partial u}{\partial x} - \gamma \frac{\partial ^2u}{\partial x^2} = h \quad , x\in[0,1], t\in(0,0.5] \label{eq:burgersfamily}
\end{equation}
with periodic BC $u(0,t)=u(1,t)$. We fix the viscosity $\gamma = 0.005$, but randomly generate the source function $h(x,t) = \sum_{j=1}^{5} A_j \text{sin}(\omega_j t + \frac{2\pi l_j x}{6} + \varphi_j)$ with the coefficients sampled from $A_j\in[-0.8,0.8]$, $\omega_j\in[-2,2]$, $l_j\in[0,1,2,3,4]$, $\varphi_j\in[-\pi,\pi]$. The problem consists of 480 PDE instances with different source terms and ICs (from $u(x,t=0) = h(x,t=0)$ and its perturbations), representing a rich diversity of dynamical processes. Each PDE instance contains 51$\times$26 spatial-temporal points.

\textbf{Hyperparameters setting.}
We fix $\lambda_{\textit{BC}} = \lambda_{\textit{IC}} = 1$ and perform a grid search to determine an appropriate $\lambda_{\textit{PDE}}$ and $\lambda_{\textit{PI}}$ for the MLP$+[Pi]^2$ and HYDRA$+[Pi]^2$ models based on the seen PDE instances, before applying them to solve the unseen PDE instances. For the PiL-PINN, we pass these hyper-parameters to the loss function and jointly learn them with the network weights during model optimization. This is an additional advantage of the pseudoinverse-in-the-loop learning. The number of nonlinear iterations for Burgers' equation (sine IC and family of IC) is 4 and 8 respectively.

\textbf{Compute environment.}
We implement our models in JAX and perform the experimental study on a workstation with Intel Xeon W-2275 Processor and 2 NVIDIA RTX 3090 GPUs.

\subsection{Results and Discussion}

Given the underlying context that data is typically sparse in many science and engineering real-world scenarios, we evaluate the performance of the various models (MLP, MLP$+[Pi]^2$, HYDRA$+[Pi]^2$, and PiL-PINN) across all 4 problems for a range of scenarios where solutions to $K$ randomly sampled tasks are available for training. The main results are in Figure~\ref{fig:main}. A representative plot of the solutions for Poisson's equation problem is given in Figures~\ref{fig:poisson}. Representative plot of the solutions obtained by the different methods for the Helmholtz equation and nonlinear Burgers' equation (Sine IC and Family of IC) are given in Figures~\ref{fig:helmholtz},~\ref{fig:sineburger} and~\ref{fig:metaburgers}.

\textbf{Improvement in Prediction Performance through MLP$+[Pi]^2$.}
In general, we note that the data-driven MLP's predictive performance is fairly poor, with relative errors ranging from 0.1 to 1, which is intuitively consistent with the small amount of training data provided and the complexity of the model physics. The predicted solutions are visibly inaccurate in Figures \ref{fig:sineburger} and \ref{fig:helmholtz}.

Excitingly, the combination of a simple data-driven training strategy and an extra psuedoinverse computation to fine-tune the final layer to better match the underlying physics-based constraints still results in substantial improvements in relative error for the Poisson's equation (2 orders of magnitude reduction) and two Burgers' equation problems (1 order of magnitude reduction) as shown in Figure \ref{fig:main}. These experiments clearly show the effectiveness of a physics-informed loss in improving the prediction of data-driven models under conditions where the available labeled data is small. The results also open the possibility for greater utilization of other data-driven models and architectures in future extensions. We note that a reason for the widespread usage of MLP architectures in PINN literature is the need for simple implementations given the complexity of PINN training. The potential extension with a physics-informed loss through fast pseudoinverse computations can greatly mitigate this issue.

In addition, the benefit from including the physics-informed loss is most pronounced when $K$ is smaller. As $K$ increases, the data-driven MLP model's relative error begins to show a decreasing trend. However, we note that the MLP$+[Pi]^2$ generally shows reasonable performance even at low $K$ in 3 of the 4 problems. Nonetheless, the data-driven MLP training does not always learn a good shared embedding for the physics-informed loss, as exemplified by the relatively higher errors in the Helmholtz problem.

\textbf{Importance of an Expressive Shared Embeddings.}
Given the importance of expressiveness of the learned embedding for the pseudoinverse computation to minimization of the physics-informed loss, we further compare the proposed neural architecture (with concatenative skip connections across the hidden layers) in the context of the multi-task learning setup. 

From Figure \ref{fig:main}, the HYDRA$+[Pi]^2$ shows significantly improved relative errors compared to the MLP$+[Pi]^2$ models. This improvement is particularly notable for the Poisson equation and the Helmholtz equation in Figure \ref{fig:helmholtz}, showing the importance of our proposed neural architecture in enabling the learning of a more expressive and performant shared embedding for the pseudoinverse computation. We also notice that the performance of HYDRA$+[Pi]^2$ (and PiL-PINN models) can be further improved by the use of an appropriate factor $F \pi$ in Eq. \ref{eq:frequency_annealing} ($F=2$ for Poisson's and $4$ for Helmholtz problems).

While the improvement was notable for the linear PDEs (Poisson's and Helmholtz equation), the reduction in the relative error for HYDRA$+[Pi]^2$ relative to the MLP$+[Pi]^2$ was much less significant for the nonlinear Burgers' equation. This is potentially due to the fact that the current zero-shot framework fundamentally only finds the best linear solution (through the pseudoinverse). 

\textbf{PiL-PINN for Nonlinear PDEs.}
As the nonlinear PDEs are currently solved in a slightly modified process through a series of iterations where the equation is linearized, a different way of learning the shared embedding where the pseudoinverse is taken explicitly into account during training can be more appropriate. Hence, we proposed the PiL-PINN methodology to explicitly handle this nuance. 

As per Figure \ref{fig:main}, the PiL-PINN shows a further reduction in relative errors for the nonlinear Burgers' problems relative to the other 3 methods (improvement relative to the MLP method is presented in Figure \ref{fig:sineburger} and Figure \ref{fig:metaburgers}). In addition, the PiL-PINN results also show a significant reduction in relative error with $K$, indicating that the proposed method indeed allows for better learning of a shared embedding for zero-shot prediction as observed for Burgers' equation.

\textbf{Computation Time.}
The training time for the PiL-PINN models range from a few seconds on the Poisson's problem to 1.5 hours, on the Burgers' (family of IC) problem ($K=240$). The PiL-PINN model takes less than 1 second to make predictions on a new PDE instance across all the problems tested. For example, it takes 54ms to make a prediction for the Burgers' equation (sine IC)\textemdash a benchmark problem frequently studied in PINN literature\textemdash at 129x51 spatial-temporal points (Figure \ref{fig:sineburger}). In contrast, others in literature have reported typical training times of $\approx$10 min to 1 hour for a wide range of canonical PDE problems \cite{wang2023expert}. This training time is also consistent with our own experience when training single PINN models to produce predictions for Burgers', Poisson, or Helmholtz equation. This means that the Pi-PINN method can produce predictions for new scenarios 100--1000$\times$ faster.

\section{Conclusions}

In this work, we demonstrate a transferable learning approach that enables rapid and accurate physics-informed solutions by incorporating a least-squares-optimal pseudoinverse computation within a closed-form head adaptation step. This approach underscores the potential of combining data-driven multi-task learning with physics-informed learning to improve PINNs’ efficiency and generalization across PDE families and parameter regimes. Even when the labeled data are limited, our results show that Pi-PINN can learn a shared, transferable embedding that supports significantly improved generalization to new PDE instances. The empirical results also suggest that architectural modifications such as concatenative skip connections are beneficial. Nonetheless, there remains substantial scope for developing neural architectures and training algorithm designs that learn transferable embeddings even more effectively, which is particularly important given the practical constraint that datasets in many scientific and engineering domains may be limited in quantity. We hope this work encourages alternative directions that move PINNs closer to robust, reusable tools for real-world applications.


\section*{Acknowledgment}
This research was in part supported by the National Research Foundation, Singapore through the AI Singapore Programme, under the project ``AI-based urban cooling technology development" (Award No. AISG3-TC-2024-014-SGKR). The authors acknowledge the use of AI tools for language editing and take full responsibility for the final manuscript.

\bibliographystyle{IEEEtran}
\bibliography{IEEEabrv,Main}

\end{document}